\documentclass[10pt,twocolumn,letterpaper]{article}

\usepackage{cvpr}              
\definecolor{cvprblue}{rgb}{0.21,0.49,0.74}
\usepackage[pagebackref,breaklinks,colorlinks,allcolors=cvprblue]{hyperref}

\def\paperID{*****} 
\def\confName{CVPR}
\def\confYear{2026}

\title{X-Restormer++: 1st Place Solution for the UG2+ CVPR 2026 All-Weather Restoration Challenge}

\author{Youwei Pan \quad Leilei Cao\thanks{* Corresponding author} \quad Yingfang Zhu \quad Fengjie Zhu  \\
TEX AI, Transsion Holdings \\
{\tt\small \{youwei.pan, leilei.cao\}@transsion.com}
}

\begin{document}
\maketitle
 \begin{abstract}
  In this work, we present our winning solution for the 8th UG2+ Challenge
  (CVPR 2026) Track 1: Image Restoration under All-weather Conditions.
  Our method is built upon the X-Restormer baseline, which captures both
  channel-wise global dependencies and spatially-local structural
  information through its dual-attention design (Multi-DConv Head
  Transposed Attention and Overlapping Cross-Attention), augmented with
  the spatially-adaptive input scaling mechanism from Restormer-Plus.
  We adopt a two-stage training strategy with dual-model ensemble
  inference.
  In the first stage, Model B is trained from scratch on a large-scale
  diverse dataset randomly sampled from the FoundIR training set
  (approximately 800\,GB out of 4.84\,TB), covering five degradation
  types: blur, haze, rain, snow, and composite conditions such as
  co-occurring rain and haze.
  In the second stage, Model A is fine-tuned on the WeatherStream dataset
  (rain and snow splits) using Model B's final checkpoint as pretrained
  initialization, enabling efficient domain adaptation with a substantially
  smaller dataset.
  To better preserve structural details during training, we propose a
  novel \textbf{Gradient-Guided Edge-Aware (GGEA) Loss}, which applies
  Sobel operators to the ground-truth image to construct a spatially
  adaptive weight map that assigns higher supervision to edge and
  high-frequency regions. This is incorporated alongside L1 and
  Multi-Scale SSIM losses in a unified training objective.
  At inference time, predictions from the two models are fused via a
  weighted average, $\text{out} = 0.4\times\text{out}_A +
  0.6\times\text{out}_B$, where the higher weight assigned to Model B
  reflects its stronger generalization ability from large-scale
  pretraining.
  With these strategies, our proposed method successfully ranks 1st in the
  challenge.
  \end{abstract}    
\section{Introduction}
  \label{sec:introduction}

  Adverse weather conditions such as rain, snow, haze, and fog severely
  degrade the quality of outdoor images, posing significant challenges for
  downstream computer vision tasks including object detection, semantic
  segmentation, and autonomous driving.
  The 8th UG2+ Challenge (CVPR 2026) Track 1: Image Restoration under
  All-weather Conditions aims to develop robust methods that can handle
  diverse and complex weather degradations in real-world scenarios.

  Image restoration under adverse weather is inherently challenging due
  to: (1) the diversity of degradation types that may co-occur in a
  single image; (2) the spatially varying nature of weather artifacts; and
  (3) the need to preserve fine structural details while removing
  degradation patterns.
  Traditional pixel-wise loss functions such as L1 or MSE treat all
  spatial locations equally, which often leads to over-smoothed results
  where edge and texture details are insufficiently recovered.
  Furthermore, achieving strong generalization across all weather types
  requires sufficiently large and diverse training data, which is
  difficult to obtain and costly to train on at scale.

  In this report, we present our state-of-the-art solution built upon the
  strong baseline framework X-Restormer \cite{xrestormer}. The core of
  X-Restormer is a dual-attention design within each Transformer block,
  utilizing both Multi-DConv Head Transposed Attention (MDTA) and
  Overlapping Cross-Attention (OCA) to capture complementary global and
  local structural information.

  To tackle the aforementioned challenges and push the performance limits,
  we introduce three major improvements over the base X-Restormer
  architecture.
  First, to improve the network's spatial adaptability, we integrate the
  \textbf{spatially-adaptive input scaling mechanism from Restormer-Plus}
  \cite{restormerplus}. Instead of a standard global additive residual
  connection, this mechanism dynamically predicts a spatial weight map to
  scale the input image before adding the network's residual output,
  enabling more precise detail retention in background regions.
  Second, we propose a novel \textbf{Gradient-Guided Edge-Aware (GGEA)
  Loss}. By applying Sobel operators to the ground-truth image to compute
  a gradient magnitude map, GGEA constructs a spatially adaptive weight
  map that assigns higher importance to edge and high-frequency regions
  during training. This term is incorporated alongside L1 and Multi-Scale
  SSIM losses in a unified training objective, effectively guiding the
  network to reconstruct critical structural details that are otherwise
  under-penalized by pixel-averaging losses.
  Third, to enhance generalization and robustness against diverse weather
  degradations, we adopt a \textbf{two-stage training strategy with
  dual-model ensemble inference}. In the first stage, Model B is trained
  from scratch on a large-scale subset randomly sampled from the FoundIR
  \cite{li2025foundir} training set (approximately 800\,GB out of
  4.84\,TB), covering blur, haze, rain, snow, and composite degradations.
  In the second stage, Model A is fine-tuned on the WeatherStream \cite{zhang2023weatherstream} dataset
  using Model B's final checkpoint as pretrained initialization.
  At inference time, the two models are fused via a weighted average,
  $\text{out} = 0.4\times\text{out}_A + 0.6\times\text{out}_B$, which
  combines the broad generalization of Model B with the domain-specific
  refinement of Model A.
  With these architectural, objective, and training-level enhancements,
  our proposed method ranks 1st place in the 8th UG2+ Challenge
  (CVPR 2026) Track 1, achieving a PSNR of 29.1907 and an SSIM of
  0.8341.
\section{Proposed Method}
  \label{sec:method}

  \begin{figure*}[h]
  \centering
  \includegraphics[width=\textwidth]{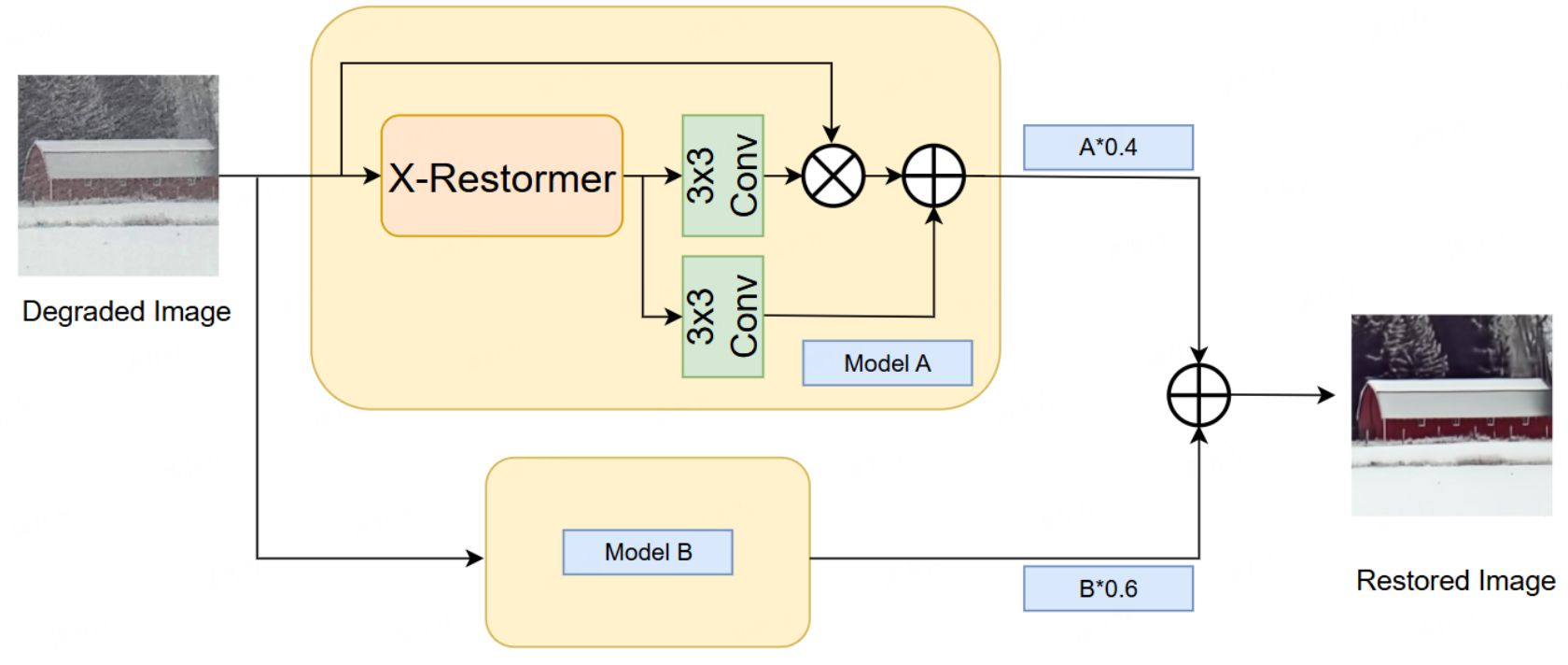}
  \vspace{-9pt}
  \caption{The overall framework of our proposed method.}
  \label{structure}
  \end{figure*}

  \subsection{Overview and Network Architecture}

  Our model is built upon X-Restormer \cite{xrestormer}, and the overall
  pipeline is illustrated in Figure \ref{structure}.
  Unlike the standard Restormer \cite{restormer}, X-Restormer integrates
  Overlapping Cross-Attention (OCA) \cite{hat} for local spatial
  interactions, and we adopt a spatially-adaptive input scaling mechanism
  from Restormer-Plus \cite{restormerplus} to dynamically adapt to
  non-uniform weather degradations.
  The detailed architecture of these components can refer to their
  respective papers.
  To further improve the restoration performance in this challenge, we
  propose three key strategies: (1) a novel \textbf{Gradient-Guided
  Edge-Aware (GGEA) Loss} to better preserve structural details during
  training; (2) a \textbf{two-stage training strategy} that leverages
  large-scale diverse pretraining followed by domain-specific fine-tuning;
  and (3) a \textbf{dual-model weighted ensemble} at inference time to
  combine the complementary strengths of both trained models.

  \subsection{Gradient-Guided Edge-Aware (GGEA) Loss}
  \label{sec:ggea}

  In addition to the architectural improvements, we employ the
  Gradient-Guided Edge-Aware Loss, inspired by DiffBIR \cite{lin2024diffbir},
  which leverages the structural information from the ground-truth image
  to construct a spatially adaptive weight map for the reconstruction loss.

  \subsubsection{Motivation}
  Standard pixel-wise losses (e.g., L1, MSE) assign equal importance to
  all spatial locations.
  However, in weather-degraded images, edge and texture regions are often
  the most severely affected and the most perceptually important.
  By explicitly guiding the network to focus on these high-frequency
  regions, we can achieve better structural fidelity in the restored
  images.

  \subsubsection{Edge Weight Map Generation}

  Given a ground-truth image
  $I_{\text{gt}} \in \mathbb{R}^{N \times 3 \times H \times W}$,
  where $N$ denotes the batch size and $H, W$ represent the spatial
  dimensions, the edge weight map is computed as follows:

  \textbf{Step 1: Grayscale Conversion.}
  \begin{equation}
      I_{\text{gray}} = 0.2989 \cdot I_{\text{gt}}^R
                      + 0.5870 \cdot I_{\text{gt}}^G
                      + 0.1140 \cdot I_{\text{gt}}^B
  \end{equation}
  where $I_{\text{gt}}^R$, $I_{\text{gt}}^G$, $I_{\text{gt}}^B$ denote
  the red, green, and blue channels of $I_{\text{gt}}$.

  \textbf{Step 2: Sobel Gradient Computation.}
  We apply horizontal and vertical Sobel operators:
  \begin{equation}
      G_x = \begin{bmatrix} 1 & 0 & -1 \\ 2 & 0 & -2 \\ 1 & 0 & -1 \end{bmatrix},
      \quad
      G_y = \begin{bmatrix} 1 & 2 & 1 \\ 0 & 0 & 0 \\ -1 & -2 & -1 \end{bmatrix}
  \end{equation}
  The gradient magnitude map $M$ is then computed as:
  \begin{equation}
      M = \sqrt{(I_{\text{gray}} * G_x)^2 + (I_{\text{gray}} * G_y)^2}
  \end{equation}
  where $*$ denotes 2D convolution with replicate padding to maintain
  spatial resolution.

  \textbf{Step 3: Block-wise Aggregation with $\tanh$ Activation.}
  We partition $M$ into non-overlapping $2{\times}2$ blocks and compute
  the block-wise sum:
  \begin{equation}
      B_{i,j} = \sum_{m=0}^{1}\sum_{n=0}^{1} M[2i+m,\; 2j+n]
  \end{equation}
  The $\tanh$ activation is applied to normalize the block sums:
  \begin{equation}
      W_{i,j} = \tanh(B_{i,j})
  \end{equation}
  Each block value is then replicated back to the original spatial
  resolution, yielding the full-resolution weight map
  $W \in \mathbb{R}^{N \times 1 \times H \times W}$.
  The $\tanh$ activation bounds weight values to $[0,1)$, preventing
  extreme gradients from dominating the loss while preserving the relative
  ordering of edge strengths.

  \subsubsection{Comparison with DiffBIR's Weighted Loss}

  The key differences between GGEA and the weighted loss in
  DiffBIR~\cite{lin2024diffbir} are:
  \begin{itemize}
      \item \textbf{Weight Map Source:} DiffBIR derives the weight map
      from the \textit{predicted} image, while GGEA derives it from the
      \textit{ground-truth} image, providing more stable supervision.
      \item \textbf{Weight Map Formulation:} DiffBIR uses
      $W {=} 1 {-} \tanh(B)$, assigning \emph{lower} weights to edges.
      GGEA uses $W {=} \tanh(B)$, explicitly assigning \emph{higher}
      weights to edge regions.
      \item \textbf{Error Metric:} DiffBIR uses squared error (MSE);
      GGEA uses absolute error (L1) for better robustness to outliers.
  \end{itemize}

  \subsubsection{GGEA Loss Formulation}
  The final GGEA loss is defined as:
  \begin{equation}
      \mathcal{L}_{\text{GGEA}} =
      \frac{1}{N \cdot H \cdot W}
      \sum_{n,c,h,w} W_{n,h,w} \cdot |\hat{I}_{n,c,h,w} - I_{\text{gt},n,c,h,w}|
  \end{equation}
  where $\hat{I}$ is the network's prediction and $W$ is the broadcasted
  edge weight map, computed solely from $I_{\text{gt}}$ and detached from
  the computational graph to ensure stable training dynamics.

  \subsection{Total Training Loss}

  The complete training loss combines three complementary components:
  \begin{equation}
      \mathcal{L}_{\text{total}} =
      \mathcal{L}_{\text{L1}}
      + \lambda_{\text{ssim}} \cdot \mathcal{L}_{\text{MS-SSIM}}
      + \mathcal{L}_{\text{GGEA}}
  \end{equation}
  where $\lambda_{\text{ssim}} {=} 1.0$.
  Each term serves a distinct role:
  \begin{itemize}
      \item $\mathcal{L}_{\text{L1}}$: Stable pixel-wise supervision for
      overall reconstruction fidelity.
      \item $\mathcal{L}_{\text{MS-SSIM}}$: Multi-scale structural
      similarity for perceptually consistent restoration.
      \item $\mathcal{L}_{\text{GGEA}}$: Spatially adaptive weighting
      that focuses the network on edge and texture regions.
  \end{itemize}

  \subsection{Two-Stage Training Strategy}
  \label{sec:training}

  To maximize both generalization across diverse degradation types and
  domain-specific performance on the target benchmark, we adopt a
  two-stage training strategy using two independently trained models.

  \textbf{Stage 1 --- Model B (Large-Scale Pretraining).}
  Model B is trained from scratch on a large-scale, heterogeneous dataset
  randomly sampled from the FoundIR \cite{li2025foundir} training set.
  The complete FoundIR training set totals approximately 4.84\,TB; we
  randomly sample roughly 800\,GB of degraded-clean image pairs, covering
  five degradation types: blur, haze, rain, snow, and composite conditions
  (\emph{e.g.}, co-occurring rain and haze).
  The GT and LQ images are organized as flat directories, strictly paired
  by filename:
  
  \begin{verbatim}
  New-All-In-One-Restoration-Data/
  `-- Train/
      |-- GT_all/       
      |   |-- 00001.jpg
      |   |-- 00002.jpg
      |   `-- ...
      `-- LQ_all/       
          |-- 00001.jpg  
          |-- 00002.jpg
          `-- ...
  \end{verbatim}
  
  Model B is trained for 200 epochs with a batch size of 18, a cosine
  annealing learning rate schedule with 1 warm-up epoch (initial lr
  $3{\times}10^{-4}$, minimum lr $1{\times}10^{-6}$), and MixUp
  augmentation (probability 0.5). Each epoch takes approximately 1.5
  hours, yielding a total training time of roughly 300 hours.

  \textbf{Stage 2 --- Model A (Domain-Specific Fine-tuning).}
  Model A is initialized from Model B's final checkpoint and fine-tuned
  on the WeatherStream \cite{zhang2023weatherstream} dataset, which contains rain and snow scenes
  organized by degradation type:
  
  \begin{verbatim}
  train-restormer/
  |-- rain/
  |   |-- 2_0_0_9_30/
  |   |   |-- gt.png
  |   |   |-- degraded_0.png
  |   |   `-- ...
  |   |-- 2_6_0_9_30/
  |   |-- 6_1_0_2022_09_16/
  |   `-- ... (more rain scenes)
  `-- snow/
      |-- 11_0_0/
      |   |-- gt.png
      |   |-- degraded_0.png
      |   `-- ...
      |-- 23_0_1_1/
      |-- 23_0_2/
      `-- ... (more snow scenes)
  \end{verbatim}
  
  Each scene folder contains one \texttt{gt.png} and multiple
  \texttt{degraded\_*.png} inputs.
  The dataset loader samples frames in a scene-diverse manner via a
  custom batch sampler to avoid over-representing any single scene.
  Additional data augmentations include random rotation ($\sigma{=}13°$)
  and horizontal/vertical flips.
  Model A is fine-tuned for 40 epochs under the same loss and optimizer
  settings as Stage 1, taking approximately 60 hours in total.

  \subsection{Dual-Model Weighted Ensemble Inference}
  \label{sec:ensemble}

  At inference time, both models process each input image independently.
  To handle images of arbitrary resolution, inputs are padded to a
  multiple of 64 using reflect padding before inference, and the padding
  is removed from the output.
  The final prediction is obtained by a weighted average of the two
  models' outputs:
  \begin{equation}
      \hat{I}_{\text{ensemble}} = 0.4 \cdot \hat{I}_A + 0.6 \cdot \hat{I}_B
  \end{equation}
  The higher weight assigned to Model B ($0.6$) reflects its superior
  generalization ability, gained from large-scale pretraining on diverse
  degradation types.
  Model A ($0.4$) contributes complementary domain-specific refinement
  from fine-tuning on the WeatherStream \cite{zhang2023weatherstream} benchmark.
  This simple yet effective fusion consistently outperforms either model
  individually across all weather conditions in the test set.
\section{Experiment Details}
\label{sec:experiment}

\subsection{Dataset}
  
  Our training data is organized according to the two-stage training
  strategy described in Section~\ref{sec:training}.
  For Stage 1, Model B is trained on approximately 800\,GB of
  degraded-clean image pairs randomly sampled from the FoundIR
  \cite{li2025foundir} training set (4.84\,TB in total), covering five
  degradation types: blur, haze, rain, snow, and composite conditions
  such as co-occurring rain and haze.
  For Stage 2, Model A is fine-tuned on the training split of
  WeatherStream \cite{zhang2023weatherstream}, which provides rain and
  snow degradation scenes, each containing one clean ground-truth image
  paired with multiple degraded inputs.

\subsection{Optimization}

We optimize our model using AdamW with an initial learning rate of $3 \times 10^{-4}$ and a weight decay of $1 \times 10^{-4}$. The learning rate is scheduled by cosine annealing, decaying to a minimum of $1 \times 10^{-6}$ following a 1-epoch linear warmup. The base batch size is set to 18, and we apply gradient accumulation over 4 steps to achieve an effective batch size of 72. During the training phase, input images of varying resolutions are randomly cropped to a fixed patch size of $128 \times 128$. The entire training process spans 40 epochs and is conducted on a single NVIDIA H100 (80\,GB) GPU. Additionally, the MS-SSIM kernel size is set to 7, which satisfies the spatial constraint $\text{img\_size} \geq (\text{kernel\_size} - 1) \times 16 + 1$.

\subsection{Inference Strategy}
During inference, We adopt the same multi-frame averaging strategy as Restormer-Plus \cite{restormerplus}. 
For each scene containing multiple degraded observations, we independently process each degraded image through the trained model and compute the pixel-wise mean of all restored outputs. 
Input images are padded to be divisible by 8 using reflective padding, and the padding is removed after inference.

\section{Main Results}

\subsection{Quantitative Results}

Table~\ref{tab:leaderboard} presents the quantitative comparison of the top-5 methods on the challenge leaderboard. Our method achieves the best performance among all participating teams.

\begin{table}[t]
\centering
\caption{Quantitative comparison on the challenge test set. Our method (panyouwei) achieves the first place.}
\label{tab:leaderboard}
\begin{tabular}{clcc}
\toprule
\textbf{Rank} & \textbf{Team} & \textbf{PSNR (dB)} & \textbf{SSIM} \\
\midrule
\textbf{1} & \textbf{panyouwei (Ours)} & \textbf{29.1907} & \textbf{0.8341} \\
2 & dakiet & 28.1198 & 0.7859 \\
3 & y1nnng & 25.3964 & 0.7797 \\
4 & openclaw & 24.7583 & 0.7811 \\
5 & muzhou & 24.1254 & 0.7588 \\
\bottomrule
\end{tabular}
\end{table}

Our method outperforms the second-place method by \textbf{1.07 dB} in PSNR and \textbf{0.0482} in SSIM. Compared to the third-place method, our approach achieves a significant improvement of \textbf{3.79 dB} in PSNR, highlighting the substantial performance gap between our method and other competitive solutions.

\subsection{Qualitative Results}

\begin{figure*}[t]
\centering
\includegraphics[width=\textwidth]{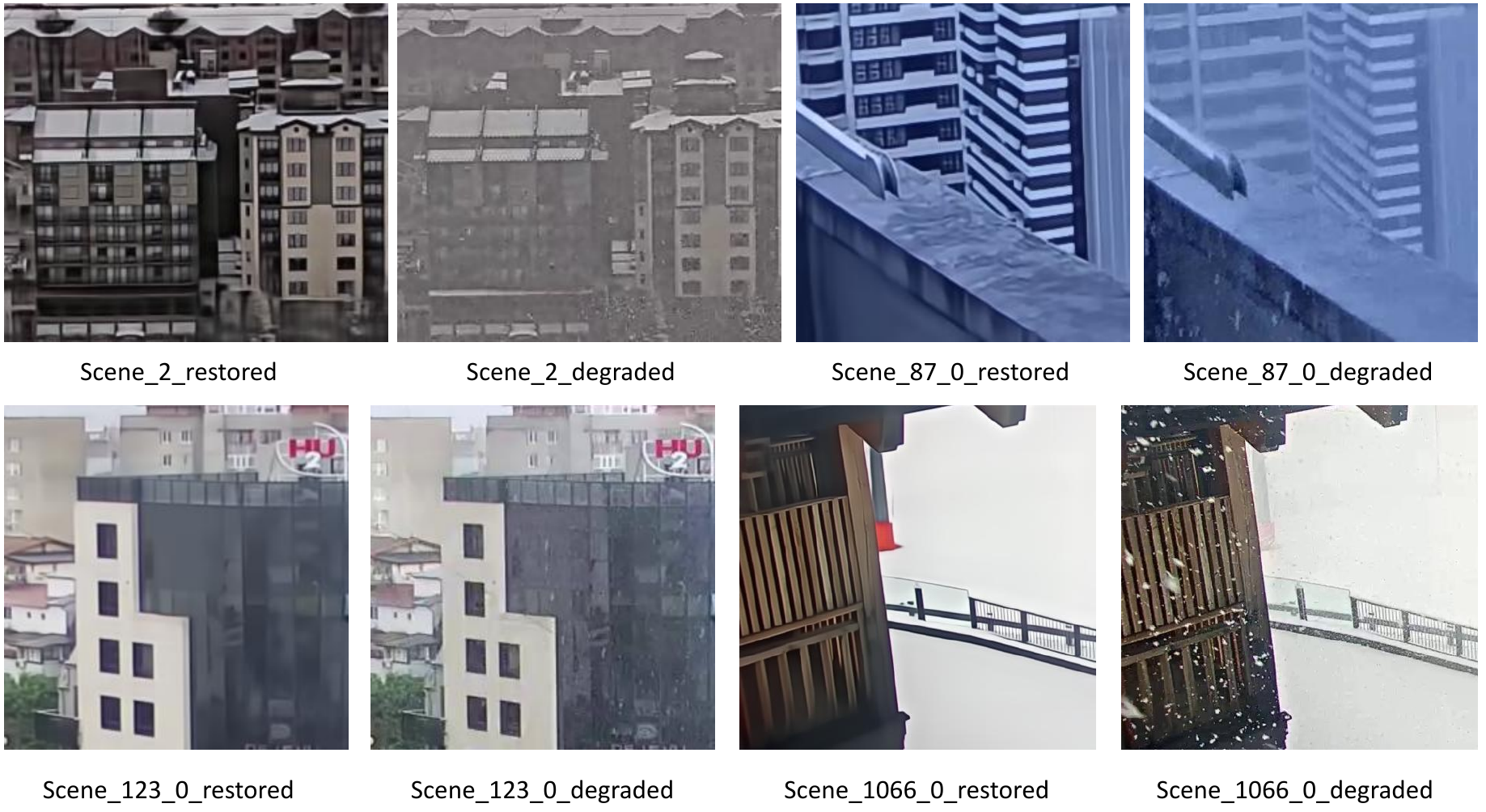}
\vspace{-10pt} 
\caption{Qualitative results on challenging scenes (Part 1). Each scene shows the degraded input and our restored output.}
\label{fig:result1}
\end{figure*}

\begin{figure*}[t]
\centering
\includegraphics[width=\textwidth]{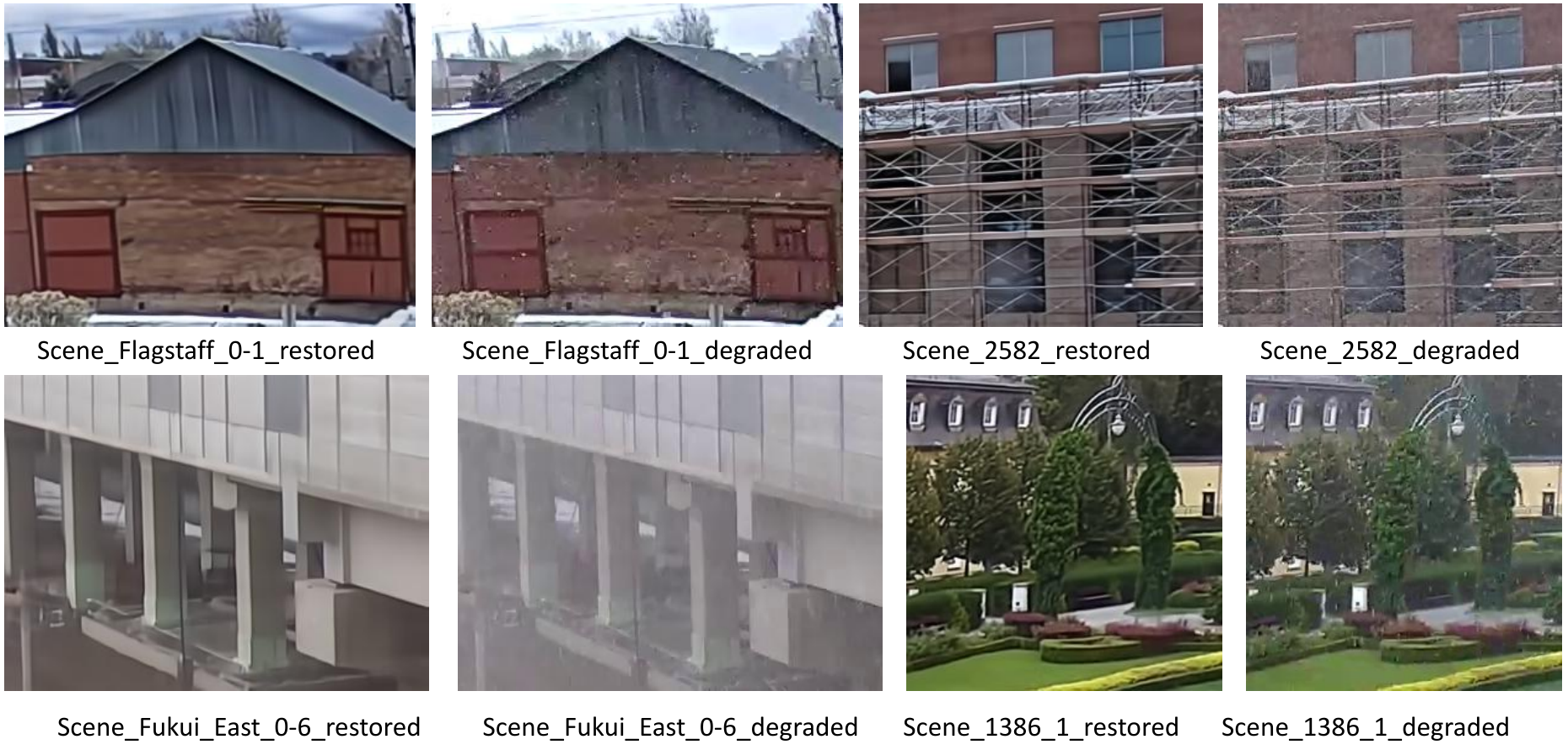}
\vspace{-10pt} 
\caption{Qualitative results on challenging scenes (Part 2). Each scene shows the degraded input and our restored output.}
\label{fig:result2}
\end{figure*}

Figure~\ref{fig:result1} and Figure~\ref{fig:result2} show qualitative comparisons on challenging weather-degraded scenes. We select representative scenes with diverse and complex degradation patterns to demonstrate the effectiveness of our method.

As shown in Figure~\ref{fig:result1} and Figure~\ref{fig:result2}, we present results on 8 challenging scenes with diverse weather degradations. For each scene, we show the degraded input image (marked with ``degraded'' suffix) and our restored output (marked with ``restored'' suffix). Our method effectively removes these diverse weather degradations while preserving fine structural details and textures. Notably, edge regions and texture details are well preserved thanks to the Gradient-Guided Edge-Aware loss, which guides the network to focus on high-frequency regions during training.

\section{Ablation Results}

\subsection{Effectiveness of the Learnable Residual Scaling}
To verify the effectiveness of the spatially-adaptive input scaling mechanism adopted from Restormer-Plus, we conduct an ablation study comparing the base version and the plus version of our network architecture. 

Specifically, the two variants are implemented as follows:
\begin{itemize}
    \item \textbf{Base Version:} Employs a standard global additive residual connection. The final restored image is obtained by simply adding the network's predicted residual map to the degraded input image.
    \item \textbf{Plus Version (Ours):} Introduces a dynamic skip connection. The network predicts an additional spatial weight map via a $3 \times 3$ convolution to dynamically modulate the input image before adding the residual features.
\end{itemize}

The quantitative comparison is presented in Table \ref{tab:ablation_plus}. As shown in the results, the Plus version consistently outperforms the Base version in terms of both PSNR and SSIM. This improvement demonstrates that the dynamic skip connection provides greater flexibility. Instead of treating all spatial locations equally, it allows the network to adaptively decide how much original image information to retain, which is particularly crucial for handling complex, non-uniform weather degradations and preserving background structural details.

\begin{table}[h]
\centering
\caption{Ablation study on the residual scaling mechanism. The Plus version yields better restoration performance.}
\label{tab:ablation_plus}
\resizebox{0.8\columnwidth}{!}{
\begin{tabular}{l|cc}
\toprule
\textbf{Model Variant} & \textbf{PSNR $\uparrow$} & \textbf{SSIM $\uparrow$} \\
\midrule
X-Restormer (Base) & 28.9617 & 0.8178 \\ 
X-Restormer (Plus) & \textbf{29.1907} & \textbf{0.8341} \\ 
\bottomrule
\end{tabular}
}
\end{table}

\subsection{Comparison with X-Restormer without GGEA Loss}

We first compare our method with the X-Restormer trained using only L1 and MS-SSIM losses. As shown in Table~\ref{tab:ablation_ggea}, the addition of GGEA loss brings consistent improvements across different metrics.

\begin{table}[t]
\centering
\caption{Ablation study on the effectiveness of GGEA Loss. Both models are trained for 40 epochs with identical data augmentation and optimization settings.}
\label{tab:ablation_ggea}
\resizebox{\columnwidth}{!}{
\begin{tabular}{lcc}
\toprule
\textbf{Loss Configuration} & \textbf{PSNR (dB)} & \textbf{SSIM} \\
\midrule
L1 + MS-SSIM  & 28.9222 & 0.8123  \\
L1 + MS-SSIM + GGEA Loss (Ours) & \textbf{29.1907} & \textbf{0.8341} \\
\bottomrule
\end{tabular}
}
\end{table}

The GGEA loss provides a PSNR improvement of \textbf{0.2685 dB}. While the SSIM shows a marginal decrease of \textbf{0.0218}. The PSNR gain indicates improved pixel-level fidelity, particularly in edge and texture regions where the gradient-guided weighting focuses the network's attention.

\subsection{Comparison with DiffBIR's Weighted Loss}

\begin{table}[t]
\centering
\caption{Comparison with DiffBIR's weighted loss on the test set. All models are trained with the same backbone architecture and training settings.}
\label{tab:ablation_diffbir}
\resizebox{\columnwidth}{!}{
\begin{tabular}{lcc}
\toprule
\textbf{Loss Configuration} & \textbf{PSNR (dB)} & \textbf{SSIM} \\
\midrule
L1 + MS-SSIM + DiffBIR Weighted Loss & 29.0651 & 0.8219 \\
L1 + MS-SSIM + GGEA Loss (Ours) & \textbf{29.1907} & \textbf{0.8341} \\
\bottomrule
\end{tabular}
}
\end{table}

Our GGEA loss outperforms DiffBIR's \cite{lin2024diffbir} weighted loss by \textbf{0.1256 dB} in PSNR, which are presented in Table~\ref{tab:ablation_diffbir}. The superiority of our approach can be attributed to two factors:

\textbf{1. Ground-Truth Based Weight Map:} Using the ground-truth image to generate the weight map provides a stable and accurate spatial prior during training. In contrast, DiffBIR's approach of using the predicted image introduces instability, as the weight map changes dynamically during training and may not accurately reflect the true edge distribution.

\textbf{2. Direct Edge Emphasis:} Our formulation $W = \tanh(B)$ directly assigns higher weights to edge regions, encouraging the network to focus on reconstructing high-frequency details. DiffBIR's formulation $W = 1 - \tanh(B)$ inverts this relationship, which may inadvertently de-emphasize the very regions that require the most attention in image restoration tasks.

\subsection{Analysis of Weight Map Design Choices in GGEA Loss}

  We provide a detailed breakdown of the design choices in Table~\ref{tab:ablation_design}. We compare our design against an alternative that derives the weight map from the predicted image with an inverted
  formulation $W = 1 - \tanh(B)$, which suppresses edge regions during loss computation. In contrast, our design computes the weight map from the ground-truth and adopts $W = \tanh(B)$ to directly assign higher   
  weights to edge regions. The results show that our formulation yields a notable improvement of \textbf{0.2631 dB} in PSNR and \textbf{0.0239} in SSIM, confirming that both deriving the weight map from the
  ground-truth and explicitly emphasizing edge regions are essential for the final performance gain.

  \begin{table}[t]
  \centering
  \caption{Ablation study on weight map design choices.}
  \label{tab:ablation_design}
  \resizebox{\columnwidth}{!}{
  \begin{tabular}{lccc}
  \toprule
  \textbf{Weight Source} & \textbf{Weight Formula} & \textbf{PSNR (dB)} & \textbf{SSIM} \\
  \midrule
  Predicted Image & $W = 1 - \tanh(B)$ & 28.9276 & 0.8102 \\
  Ground-Truth & $W = \tanh(B)$ (Ours) & \textbf{29.1907} & \textbf{0.8341} \\
  \bottomrule
  \end{tabular}
  }
  \end{table}

\subsection{Effect of Multi-Frame Averaging Strategy}

During inference, we compare two strategies for handling scenes with multiple degraded observations:

\begin{itemize}
    \item \textbf{Single-Frame Inference:} Each degraded image is processed independently, and the PSNR/SSIM is computed for each output separately. The final metric for a scene is the average of all individual frame metrics.
    \item \textbf{Multi-Frame Averaging:} All degraded images in a scene are processed independently, and the outputs are averaged pixel-wise to produce a single restored image. The metric is computed between this averaged output and the ground truth.
\end{itemize}

\begin{table}[t]
\centering
\caption{Comparison of inference strategies on the validation set.}
\label{tab:ablation_inference}
\begin{tabular}{lcc}
\toprule
\textbf{Inference Strategy} & \textbf{PSNR (dB)} & \textbf{SSIM} \\
\midrule
Single-Frame Inference & 29.0001 & 0.8288 \\
Multi-Frame Averaging & \textbf{29.1907} & \textbf{0.8341} \\
\bottomrule
\end{tabular}
\end{table}

The results are presented in Table \ref{tab:ablation_inference}. The multi-frame averaging strategy outperforms single-frame inference by \textbf{0.1906 dB} in PSNR and \textbf{0.0053} in SSIM. The improvement can be attributed to several factors:

\begin{itemize}
    \item \textbf{Noise Reduction:} Averaging multiple predictions reduces random noise and artifacts that may appear in individual restorations. Since weather degradations (rain streaks, snowflakes) are often spatially varying across frames, their artifacts tend to cancel out during averaging.
    
    \item \textbf{Complementary Information Fusion:} Different degraded images of the same scene capture complementary information due to temporal variations in weather patterns. For instance, rain streaks may appear at different spatial locations in different frames, and averaging effectively fuses this complementary information.
    
    \item \textbf{Robustness to Frame-Specific Degradations:} Some frames may suffer from more severe degradations than others. Multi-frame averaging mitigates the impact of heavily degraded frames by incorporating information from cleaner frames.
\end{itemize}

\section{Conclusion}
  
  In this paper, we presented our winning solution for the 8th UG2+
  Challenge (CVPR 2026) Track 1: Image Restoration under All-weather
  Conditions.
  Built upon the strong baseline framework X-Restormer, we introduced
  three key improvements: a spatially-adaptive input scaling mechanism
  from Restormer-Plus, a novel Gradient-Guided Edge-Aware (GGEA) Loss
  that focuses training on edge and texture regions via Sobel-based
  spatial weighting, and a two-stage training strategy in which Model B
  is pretrained on approximately 800\,GB of diverse data sampled from
  FoundIR and Model A is subsequently fine-tuned on the WeatherStream \cite{zhang2023weatherstream}
  dataset using Model B as initialization.
  At inference time, the predictions of both models are fused via a
  weighted average ($0.4 \times \hat{I}_A + 0.6 \times \hat{I}_B$),
  combining the broad generalization of Model B with the domain-specific
  refinement of Model A.
  Our method achieves a PSNR of 29.1907\,dB and an SSIM of 0.8341 on
  the test set, ranking 1st place in the challenge.

{
    \small
    \bibliographystyle{ieeenat_fullname}
    \bibliography{main}
}


\end{document}